\documentclass[runningheads]{llncs}
\usepackage{graphicx}
\begin{document}
\title{Semantic Labeling of Human Action  For Visually Impaired And Blind People Scene Interaction}
%
%
\author{Leyla Benhamida\inst{1} \and
Slimane Larabi\inst{1} }
\authorrunning{L.Benhamida et al.}

\institute{RIIMA Laboratory, Computer Science Faculty\\ USTHB University, BP 32 EL ALIA, 16111, Algeria \\
\email{{lbenhamida,slarabi}@usthb.dz}\\
}
\maketitle              
\begin{abstract}
The aim of this work is to contribute to the development of a tactile device for visually impaired and blind persons in order to let them to understand actions of the surrounding people and to interact with them. First, based on the state-of-the-art methods of human action recognition from $RGB-D$ sequences, we use the skeleton information provided by Kinect, with the disentangled and unified multi-scale Graph Convolutional $(MS-G3D)$ model to recognize the performed actions. We tested this model on real scenes and found some of constraints and limitations. Next, we apply a fusion between skeleton modality with $MS-G3D$ and depth modality with CNN in order to bypass the discussed limitations. Third, the recognized actions are labeled semantically and will be mapped into an output device perceivable by the touch sense.

\keywords{Human Action Recognition  \and RGB-D images \and Deep Learning \and Graph Convolutional Neural Network \and Semantic labeling}
\end{abstract}

\section{Introduction}

Humans are in interaction with the surrounding environment all the time in order to accomplish daily activities. The interactions are based on scene understanding which represents a complicated task for the visually impaired and blind people. They face a range of practical difficulties due to the absence of the seeing sense, which is crucial in order to understand the surrounding scene and be able to interact with it.\\
Recently, many aid systems based on computer vision and image processing algorithms have been introduced and developed to meet the visually impaired and blind people's needs with scene understanding and interaction. However, most of the proposed aid systems, even traditional ones such as white canes and dog assistants, have been provided to facilitate navigation \cite{1,2,3,4}, obstacle avoidance, localisations, and grasping objects \cite{5}, without considering interactions with the surrounding people which represent a huge part of our daily activities. visually impaired and blind people needs to have an impression of human actions in order to be able to interact with them and get along with society.  Such an aid system is divided into two tasks: first, human action recognition, and then, semantic labeling to represent the recognized action on an output device.  \\

Human action recognition is a challenging topic even the large amount of papers found in the literature. Many methods and approaches have emerged in the last few years. The earliest approaches were based on RGB videos \cite{6,7} which gave promising results, but their recognition accuracy is still relatively low, even when the scene is free of clutter.  It is hard to segment the human body because of several influencing factors such as complex background, clothing color, and illumination variation.\\

After the release of depth sensors, many recent human action recognition research papers have considered involving the three-dimension data in a form of RGB-D images \cite{8,9}. There are more robust to background noises, illumination and color variations. This eases the segmentation of the human body and allows action recognition researchers to focus their effort more on getting robust feature descriptors to describe the actions rather than on low-level segmentation.\\

\begin{figure}
\begin{center}
\includegraphics[height=0.23\textwidth, width=\textwidth]{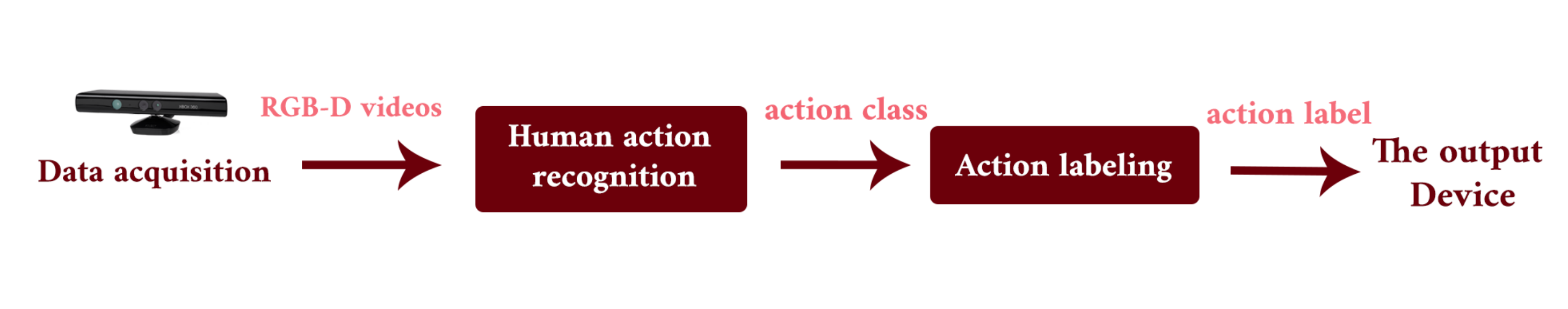}
\end{center}
\caption{Visually impaired aid system for human actions understanding. It receives RGB-D sequence as input, recognizes the performed actions and transmits the corresponding semantic labels to the output device.}
\label{fig1}
\end{figure}

In this work, we propose an aid system that enables visually impaired and blind people to have an impression of human actions and be able to interact with them. It includes two tasks: $(1)$ human actions recognition based on RGB-D videos and $(2)$ Semantic labeling to indicate the recognized action class that would be transmitted and represented on an output device (see Fig.~\ref{fig1}).

The rest of this paper is organized as follows: Section $2$ is devoted to related works in human action recognition. In section $3$, we present the used disentangled unified multi-scale graph convolutional and the results obtained after testing the model on real scenes. In section $4$, we propose our solution to tackle the discussed limitations. Finally, the action semantic labeling is described in section $5$. We conclude this paper in section $6$ and present the future works.

\section{Related works}{
Kinect-based HAR systems can be broadly classified into three categories based on the type of input data: skeleton-based, depth-based, and hybrid systems.

\subsection{Depth-based approach}{
It is based on extracting information from depth maps and exploring different points. Most existing depth-video-based action recognition methods use global features such as space-time volume information and silhouettes. For instance, in \cite{10} a histogram of oriented 4D normals (HON4D) is introduced as a Spatio-temporal depth video representation by extending the histogram of oriented 3D normals to 4D by adding the time derivative. Some other methods \cite{11,12} use local features where a set of interest points are detected. Then, depth features are extracted from the local neighborhood of each interest point. Apart from these mentioned methods which use handcrafted features, we find deep learning based methods \cite{13,14,15,16} which have attracted the interest of researchers lately. For example, Wang et al.\cite{13} trained a 3-channel deep Convolutional Neural Network (CNN) on Hierarchical Depth Motion Maps (HDMMs) that were built to extract the motion information and body shape. In \cite{14}, multi-stream deep neural networks were used to learn the semantic relations among action attributes.
}

\subsection{Skeleton-based approach}{
It explores the skeleton information which is represented by 3D positions of human key joints. Recently, this modality has attracted much attention in the research community. One advantage of using skeleton information over the other modalities and RGB sequence is that the extracted skeleton-based representation of human movement is very compact. This significantly reduces the computational cost of the recognition task. In addition, it is view-invariant and shows better recognition performance with noisy backgrounds. Earlier skeleton-based methods were using hand-crafted features. In \cite{17} a new type of feature, called EigenJoints, was proposed to combine action information, including static postures, motion, and overall dynamics. In \cite{18}, they proposed to represent joint locations by a histogram of the 3D joint positions called HOJ3D from which they collect vectors to build posture words. Then, a hidden Markov model is trained on those posture words to classify actions.\\
Recently, a lot of deep learning methods on skeleton information have been introduced and they are still on the rise. In \cite{19}, a spatial-temporal Long Short Term Memory (LSTM) was presented to learn both spatial and temporal relationships among joints. CNN-based methods usually create a pseudo-image \cite{20} using the joint coordinates and learn a feature map to achieve action classification. Recurrent Neural Networks (RNN) methods on the other hand are simply trained with a sequence of joint-coordinate vectors to infer the action \cite{21}. However, this way of using the joint coordinates misses the robustness in learning that can be achieved by an interpretation of the relationships between two connected joints and the overall structure of the skeleton.\\
From another perspective, the human skeleton takes after a topological graph. It can be restated as a graph with joint connections as edges and the joint themselves as nodes. The nodes would contain position information (3D coordinates). Furthermore, temporal edges can be added to connect joint nodes in one frame (i) to their adjacent nodes in the next frame (i+1) (Fig.~\ref{fig2}).
\begin{figure}
\begin{center}
\includegraphics[width=0.4\textwidth]{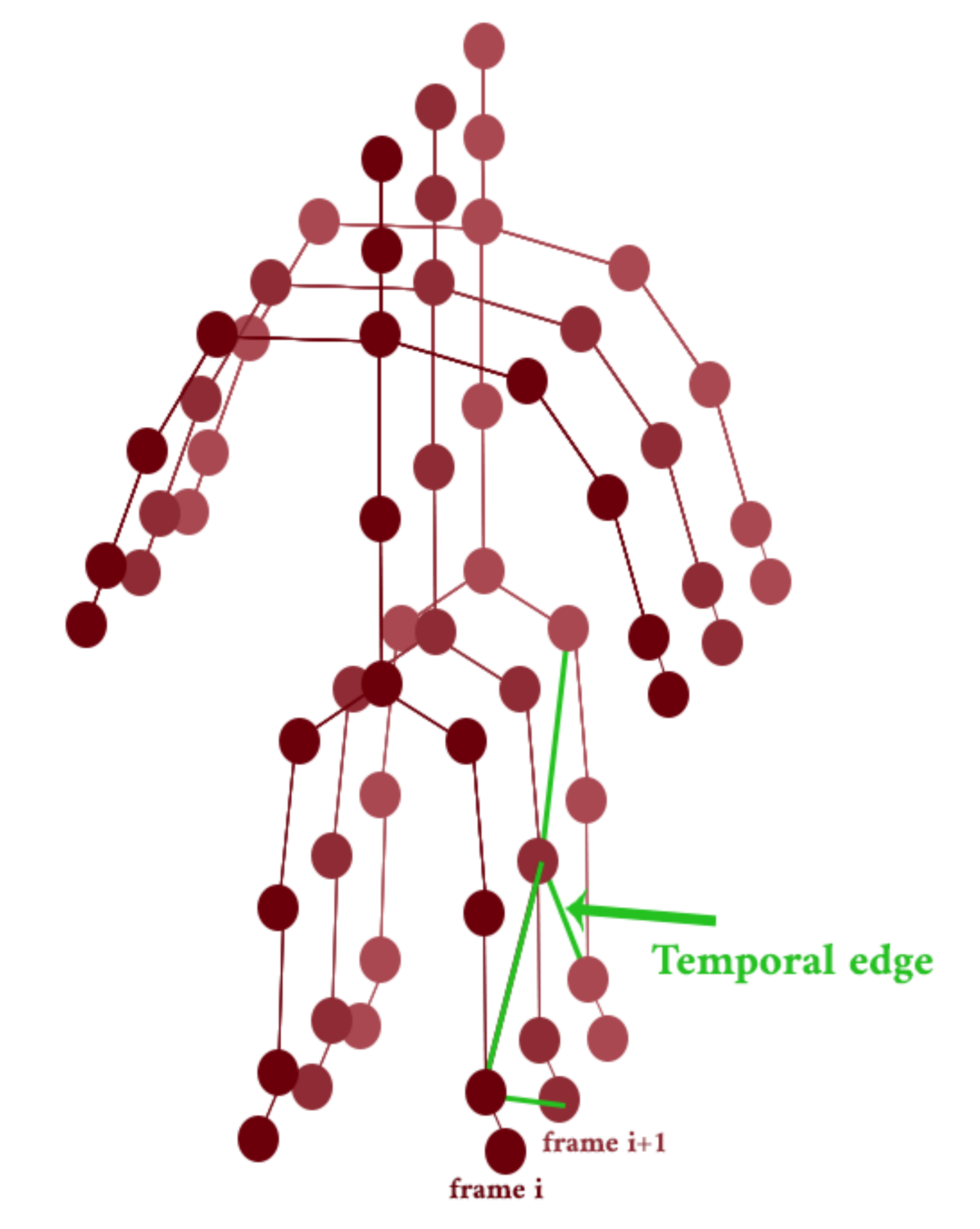}
\end{center}
\caption{The spatio-temporal representation of a skeleton.} \label{fig2}
\end{figure}

Therefore, the task of skeleton-based action recognition has also been addressed using Graph Convolution Networks(GCN). GCN treats the body skeleton as a graph, instead of sequence vector or matrix image. Recently a great number of excellent GCN-based action recognition techniques have emerged. Yan et al. \cite{22} first proposed an ST-GCN network to model the skeleton data. Their method involves successive spatial graphs and one-dimensional temporal graph convolution blocks. An adjacency matrix and a feature map of a Spatio-temporal graph are injected into the input layer of the ST-GCN. This new approach achieved state-of-the-art performance when tested on benchmark datasets. As a result, numerous ST-GCN variants were proposed within the past few years, tackling specific limitations existing in the original implementation.
}
\subsection{Hybrid approach}
It combines the two input data types (skeleton+ depth data)\cite{23,24,25}. This approach helps when there are interactions with subjects or when the actions have similar motion trajectories. For example, Chaaraoui et al. \cite{25} proposed to fuse the skeleton and silhouette shape features.\\ \\ \\

In this paper, we present a method that falls into the second approach which uses a GCN classifier. Since GCN has achieved considerable success in Human Action Recognition over a short period, the research trend is going to significantly shift from CNN and RNN methods toward GCN methods, and a lot of potential future works are underway to further investigate GCN. This was one of the reasons that led us to choose to exploit the disentangled unifying multi-scale GCN (MS-G3D)\cite{26} which represents one of the most powerful models in skeleton-based human action recognition. It should be noted that in this research, Human action recognition is not our main focus, we have just used methods from the state of the art.\\

}

\section{The disentangled unified multi-scale GCN for human action recognition}

This model is based on a powerful disentangled multi-scale aggregation scheme that leads to effectively capture wide joint relationships on human skeleton. This aggregation also addressed the biased weighting problem in existing approaches. In addition, it uses a unified spatial-temporal graph convolution operator which facilitates direct information flow across space-time for effective feature learning (G3D). Integrating the disentangled aggregation scheme with G3D gives a powerful feature extractor (MS-G3D) with multi-scale receptive fields across both spatial and temporal dimensions. The direct multi-scale aggregation of features in space-time further boosts the model performance.

\subsection{Experiments and test}
\begin{figure}
\begin{center}
\includegraphics[width=0.45\textwidth]{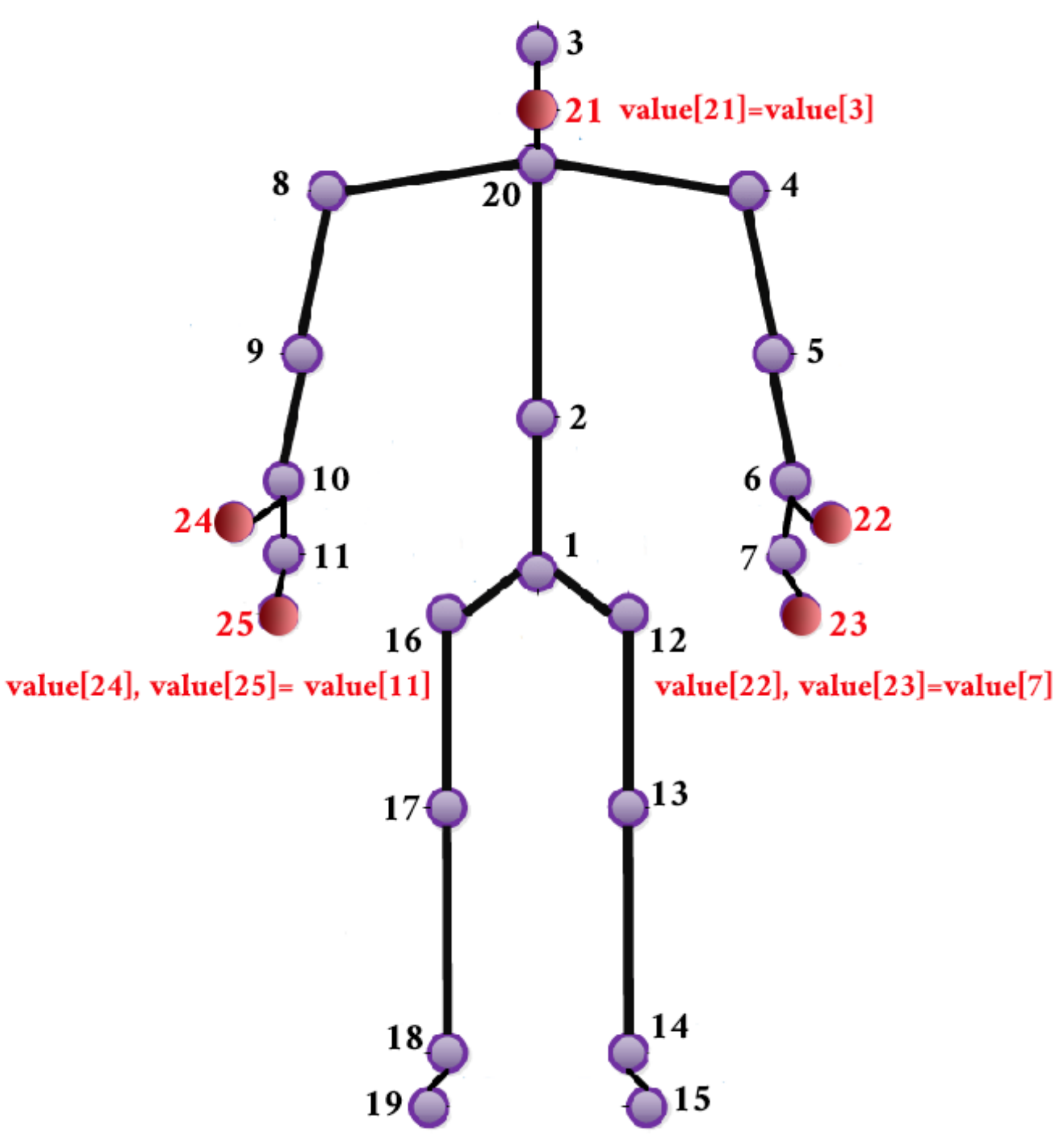}
\end{center}
\caption{Skeleton joints provided by Kinect v1. the joints that are in red are provided by kinect v2 only. We gave joints number 24 and 25 same coordinates as joint 11, and joints 22 and 23 the coordinates of joint 7, and joint 21 coordinates of joint 3.} \label{fig3}
\end{figure}

The pre-trained MS-G3D model has been trained on NTU RGB+D 60 \cite{27} dataset which is so far the largest action dataset. It contains 56,880 video sequences performed by 40 subjects and over 4 million frames with 60 classes. They have been captured from 80 views with Kinect v2. It is a very challenging dataset due to the large intra-class, sequence lengths and viewpoints variations. It has two standard evaluation protocols including cross-subject (x-sub) and cross-view(x-view). The model has achieved recognition accuracy of 91,5\% with the X-Sub protocol and 96,2\% with the X-View protocol. It significantly outperforms the state of the art methods.\\
We used Kinect v1 to capture the human body skeleton. It provides 20 3D-positions of body joints. Since the input layer of the used MS-G3D has 25 inputs that represent the 25 body joints provided by Kinect v2, we gave the other 5 joints (21, 22,23, 24 and 25), that are not provided by Kinect v1, the values of the neighboring joints (see
Fig.~\ref{fig3}).\\

Then, we preprocessed the captured data by applying normalization and translation following \cite{28,29}.
In our first attempt, we focused on some actions during the test which are: Drink, brush teeth, brush hair, eat snacks/meal, phone call, playing with/using phone, throw, sit down, stand up, clapping, hand waving, jump up, kicking something, falling, fan self/feeling warm, writing, reading, put on shoes/take off shoes and wipe face. We tested the MS-G3D model on real scenes, in real-time, where each action has been performed ten times from different viewpoints by one person. The results are shown in Table.~\ref{tab1}.
\begin{table}
\caption{Score of each action after being performed  ten times from different view-points.}\label{tab1}
\begin{center}
\renewcommand{\arraystretch}{1}
\begin{tabular}{cccc}
\hline
\large\bfseries Action  & \large\bfseries Score  & \large\bfseries Confused with & \\
\hline
 Drink &  10/10 & &  \\
\hline
Brush teeth &  6/10 & phone call &  \\
\hline
Brush hair & 8/10   &  touch neck & \\
& & phone call &  \\
\hline
Eat meal/snacks & 10/10  & & \\
\hline
Throw & 7/10 & salute &  \\
\hline
Sit down & 5/10  & take off/put on shoes & \\
\hline
Stand up & 8/10 & head-shack&  \\
\hline
Jump up & 9/10 &  & \\
\hline
Clapping & 2/10 & writing & \\
& & reading &  \\
& & playing with/using phone &  \\
\hline
Hand waving & 6/10 & fan self &  \\
\hline
Playing with/using phone & 8/10 & writing & \\
\hline
Phone call & 8/10 & brush teeth & \\
\hline
Kicking something & 10/10 &  & \\
\hline
Falling & 5/10 & take off/put on shoes & \\
\hline
Fan self & 10/10 &  & \\
\hline
Writing & 5/10 & reading & \\
& & playing with/using phone & \\
\hline
reading & 6/10 & writing & \\
& & playing with/using phone & \\
\hline
Take off/put on shoes & 10/10 &  & \\
\hline
\end{tabular}
\end{center}
\end{table}

\subsection{Discussion}
From Table 1., we can notice large confusions among some actions. We can distinguish three cases that cause those confusions. \textbf{1)First case:}  with actions that have very similar skeleton sequences and modest movement differences, like the action throw and the action salute. In this case, the performance of the recognition depends on the precision of the captured joint positions. \textbf{2)Second case:}  with actions of type human-object interaction that have very similar motion trajectories. In order to differentiate them, we need to collect more features about the used object, for example, reading, playing/using phone and writing. \textbf{3)Third case:} with actions that involve fingers and hands. For instance, the model failed at recognizing the action 'clapping' because of the lack of information about the hand form. This lack is due to the absence of the 5 joints while using Kinect v1 instead of Kinect v2.\\
In conclusion, we can say that there are two main factors that have a large impact on the MS-G3D recognition performance. The first factor is the Kinect and its precision of the captured data. The MS-G3D model is relatively weak in distinguishing the actions when skeleton joints are not very well captured and cluttered. The second factor is the amount of information that can be extracted from the skeleton. It is not sufficient to recognize some actions that require details about specific body parts as hands, or about the involved object in case of human-object interaction.\\
To overcome this last problem, we propose to exploit the depth modality in order to get more information and features about body parts and the used object.  The depth modality contains important information such as silhouette and texture of both body and object which will help with human-object interactions and with actions that have very similar skeleton motion trajectories. The fusion of these features with the skeleton features extracted by the MS-G3D can boost the recognition performance.
\section{GCN and depth modality fusion:}
Currently, we are working on a new action recognition framework. We aim to fuse the two types of data sequence: skeleton information with the MS-G3D mentioned above, and depth maps which will be transformed into a descriptor that assembles the input sequence into one image namely Depth Motion Image (DMI). This last descriptor will be fed to a CNN. Then, the prediction scores provided by both MS-G3D and CNN will be combined using a score fusion operation to get a high score of the correct action.\\
\textbf{Depth Motion Image:} it provides a description of the overall action appearance by accumulating all depth maps of the action overtime to generate a uniform representation. It captures the changes in depth of the moving body parts. It also captures the silhouette of the body and the object involved. The DMI representation provides distinctive features for each action which ease the feature extraction task for the CNN model. It is calculated as follows:

\begin{equation}
DMI(i,j)=255-min(I(i,j,t)) \forall t \in [k, k+N-1]
\end{equation}

where I(i, j, t) is the pixel position (i, j) of the frame I at time t,  and N represents the total number of frames. The pixel value of the DMI image is the minimum value of the same pixel's position of the depth maps sequence. The resulting image is normalized by dividing each pixel value by the maximum value of all pixels in the image, then the region of interest (ROI) is cropped to get rid of uninformative black pixels.
\section{Action semantic labeling}
\begin{figure}
\centering
\begin{minipage}{.5\textwidth}
  \centering
  \includegraphics[width=.5\linewidth]{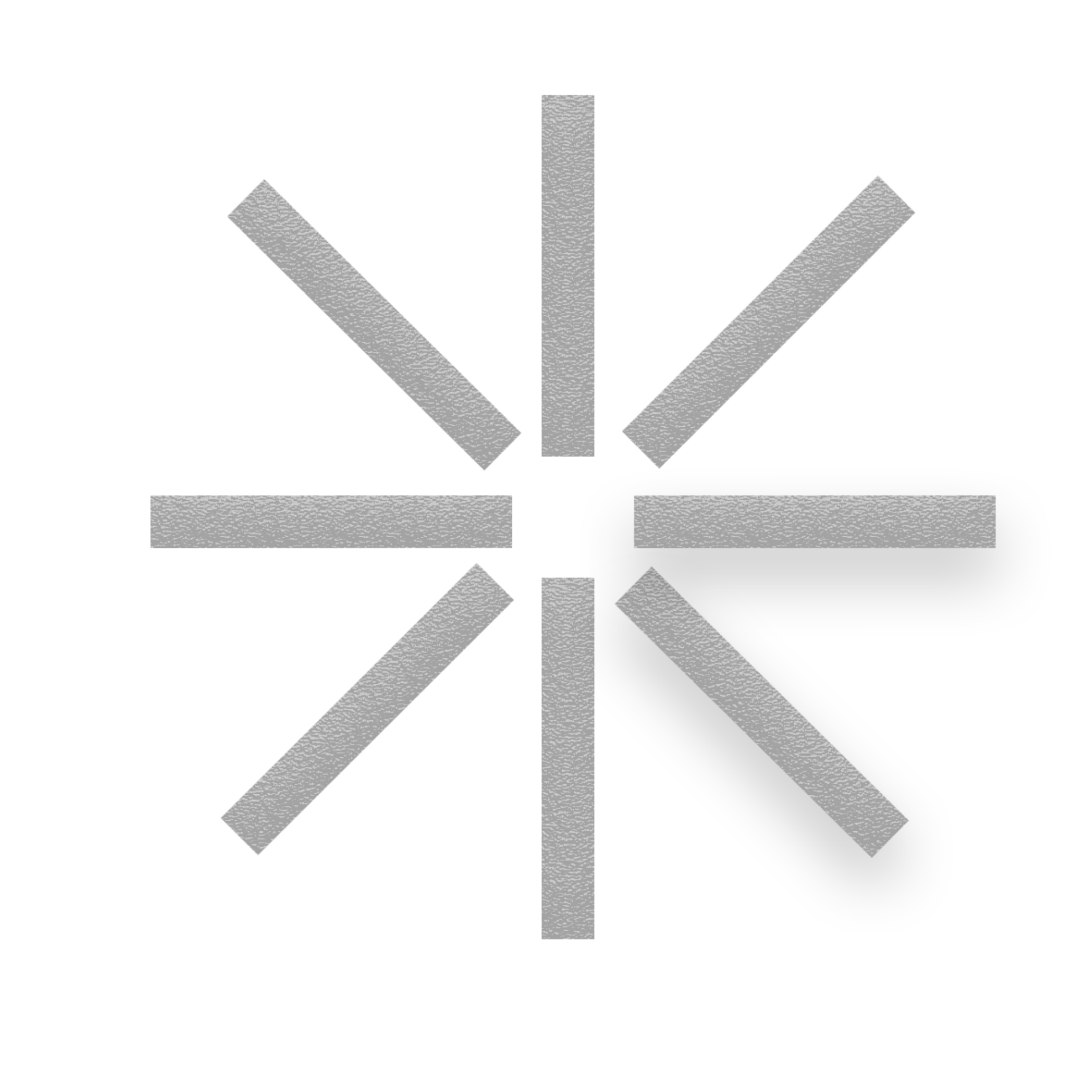}
  \caption{One node.}
  \label{fig4}
\end{minipage}%
\begin{minipage}{.5\textwidth}
  \centering
  \includegraphics[width=.5\linewidth]{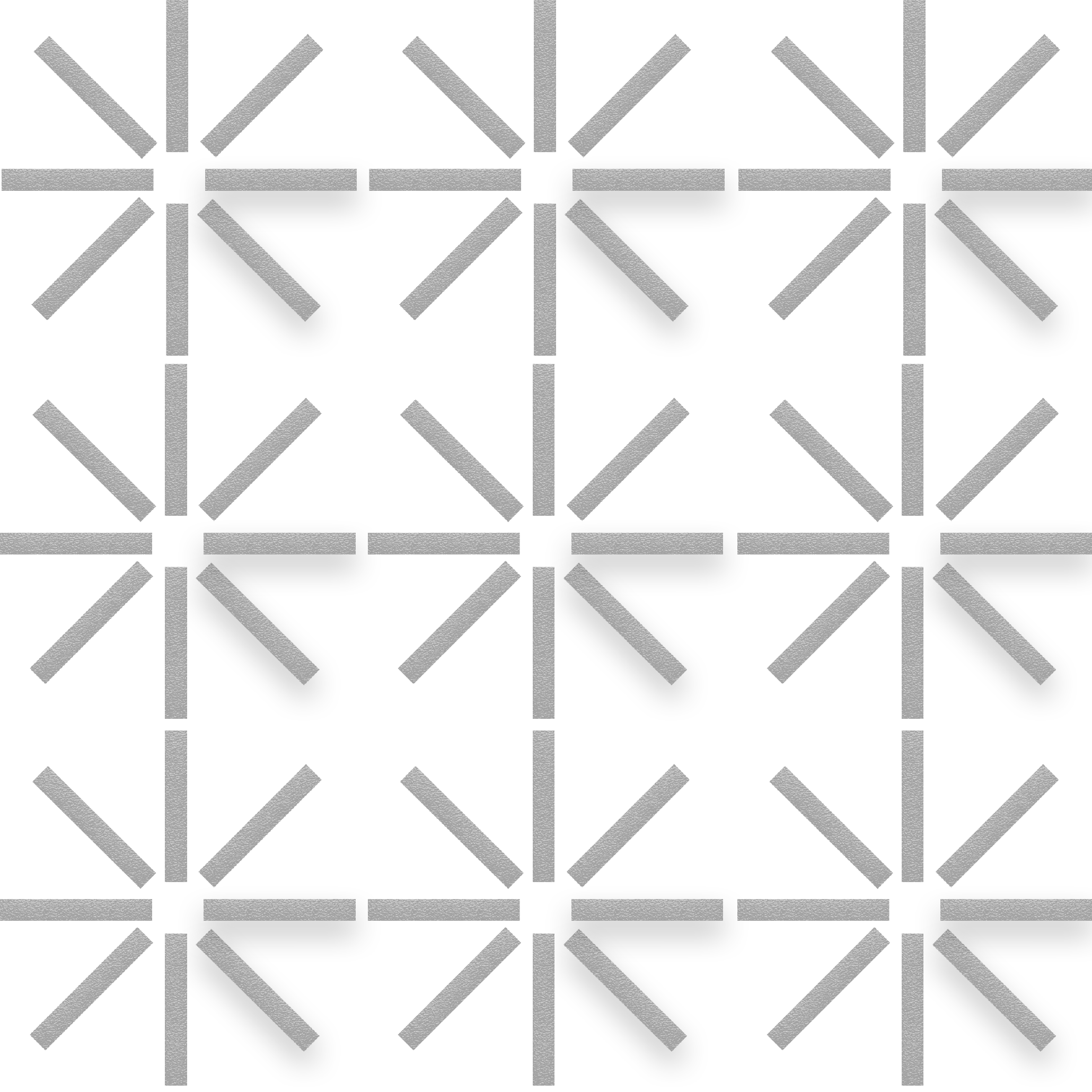}
  \caption{One cell of 9 nodes.}
  \label{fig5}
\end{minipage}
\end{figure}

We were inspired by the skeleton representation in order to elaborate a novel action semantic coding. Since skeletons are represented by lines (edges) connecting different joints, we choose lines to represent action classes. Moreover, lines are the pioneering forms to draw or to represent anything.  they are the most important shapes in the process of drawing for their flexibility, as all geometric shapes can be formed by continuous series of lines. In addition, we followed the same device design concept in \cite{30,31} that was proposed to represent objects. The device is explored by the touch where different objects are represented by raised cylinders. In our case, instead of using cylinders, we use cuboids. The lines will be provided by raised cuboids on the device. They are placed in a circular way to form nodes as shown in Fig.~\ref{fig4}.\\

Each action class is represented with a cell of 9 nodes (3x3). The head of the body is provided by raising one full node. The illustrative labeling is derived from the skeleton form when the action is performed. This coding is simple, easy to deal with, and can represent an important number of action classes just by adjusting lines.\\

\begin{table}
\caption{Some of our proposed action semantic labels. Each action is represented by raised cuboids. Black lines are the raised cuboids.}\label{tab2}
\begin{center}
\renewcommand{\arraystretch}{3}
\begin{tabular}{ccc}
\hline
\bfseries Action class & \bfseries Label  & \bfseries Labeling on the device\\
\hline
 Clapping &  {\includegraphics[width=0.18\textwidth]{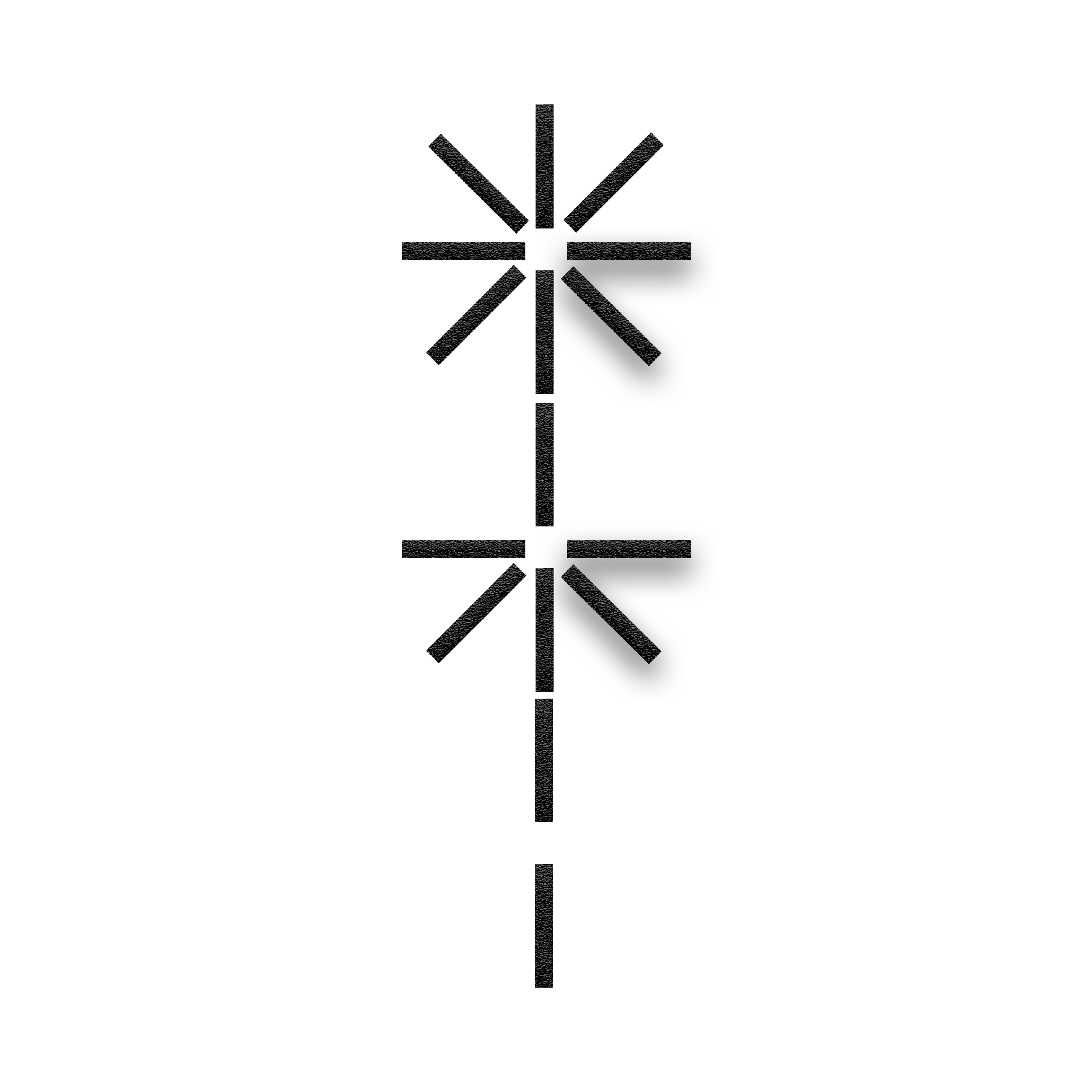}} & {\includegraphics[width=0.18\textwidth]{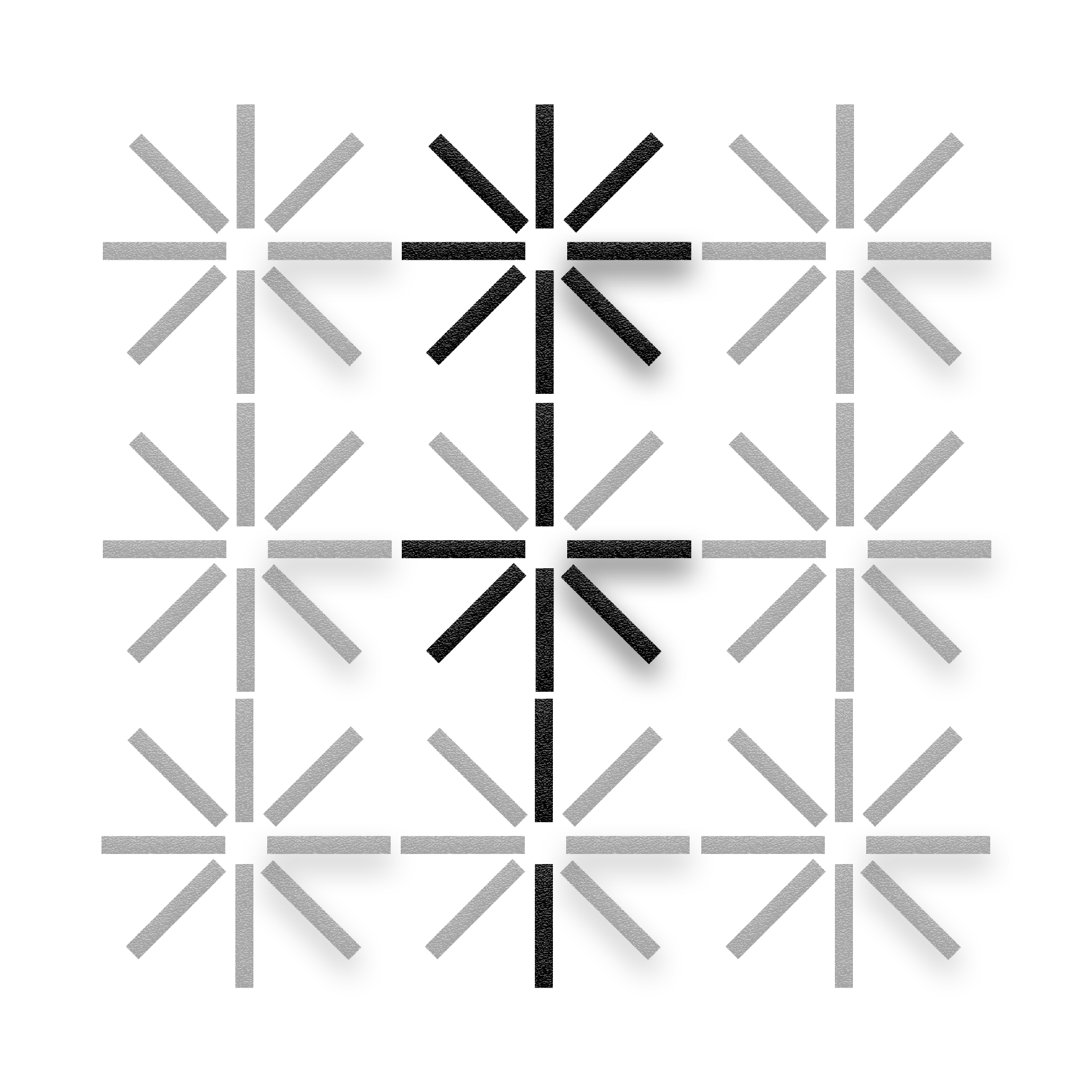}} \\
\hline
Phone call &  {\includegraphics[width=0.18\textwidth]{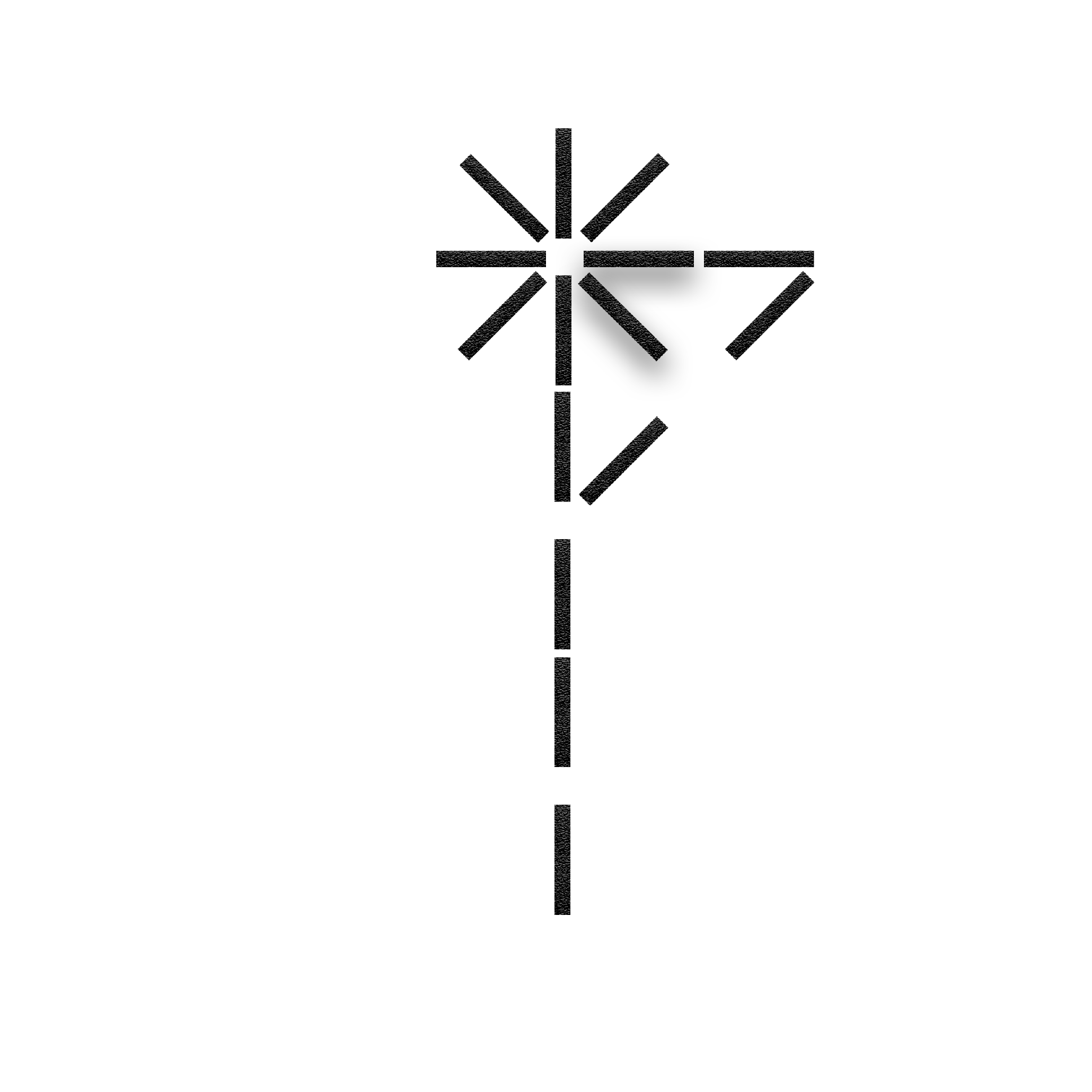}} & {\includegraphics[width=0.18\textwidth]{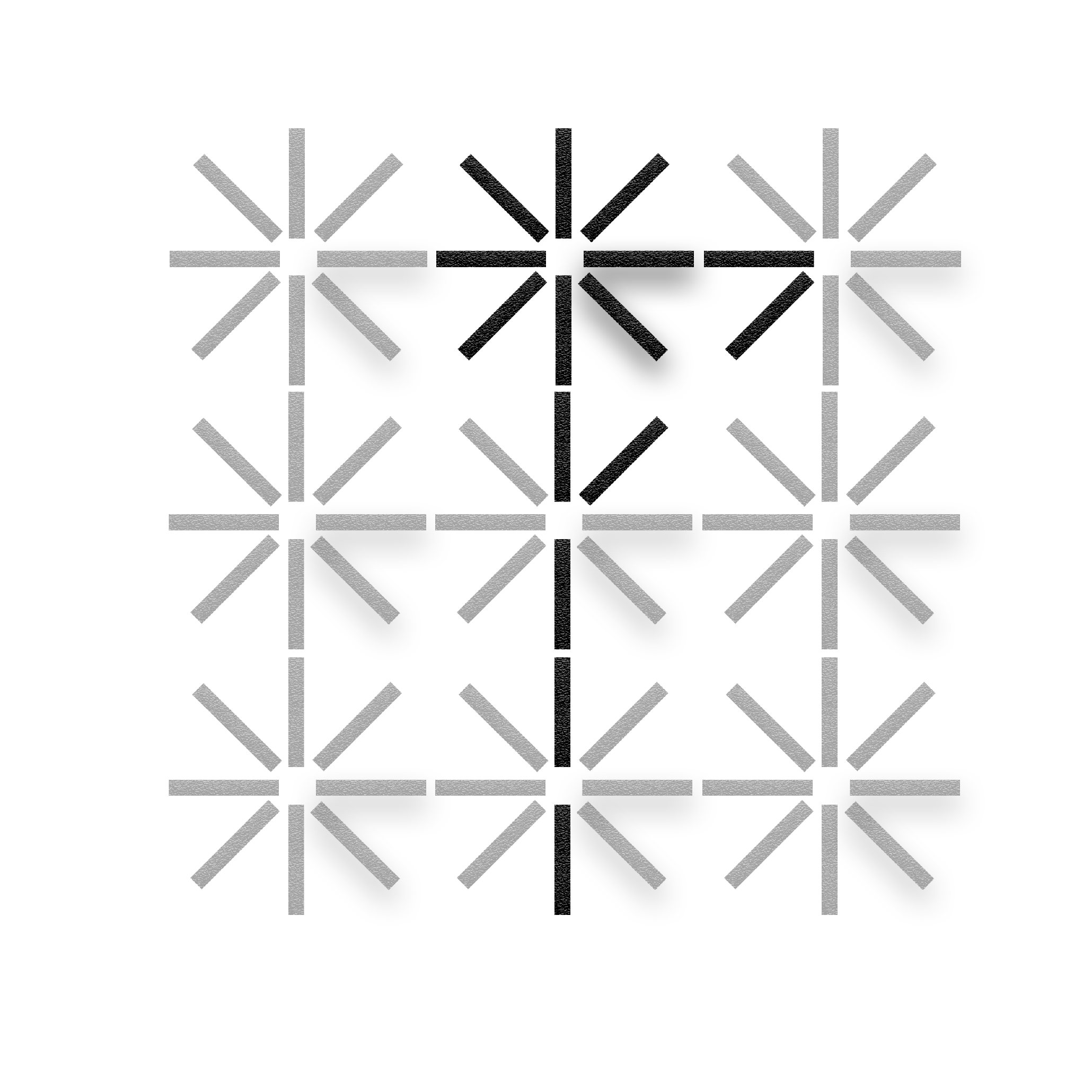}}\\
\hline
Drinking &  {\includegraphics[width=0.18\textwidth]{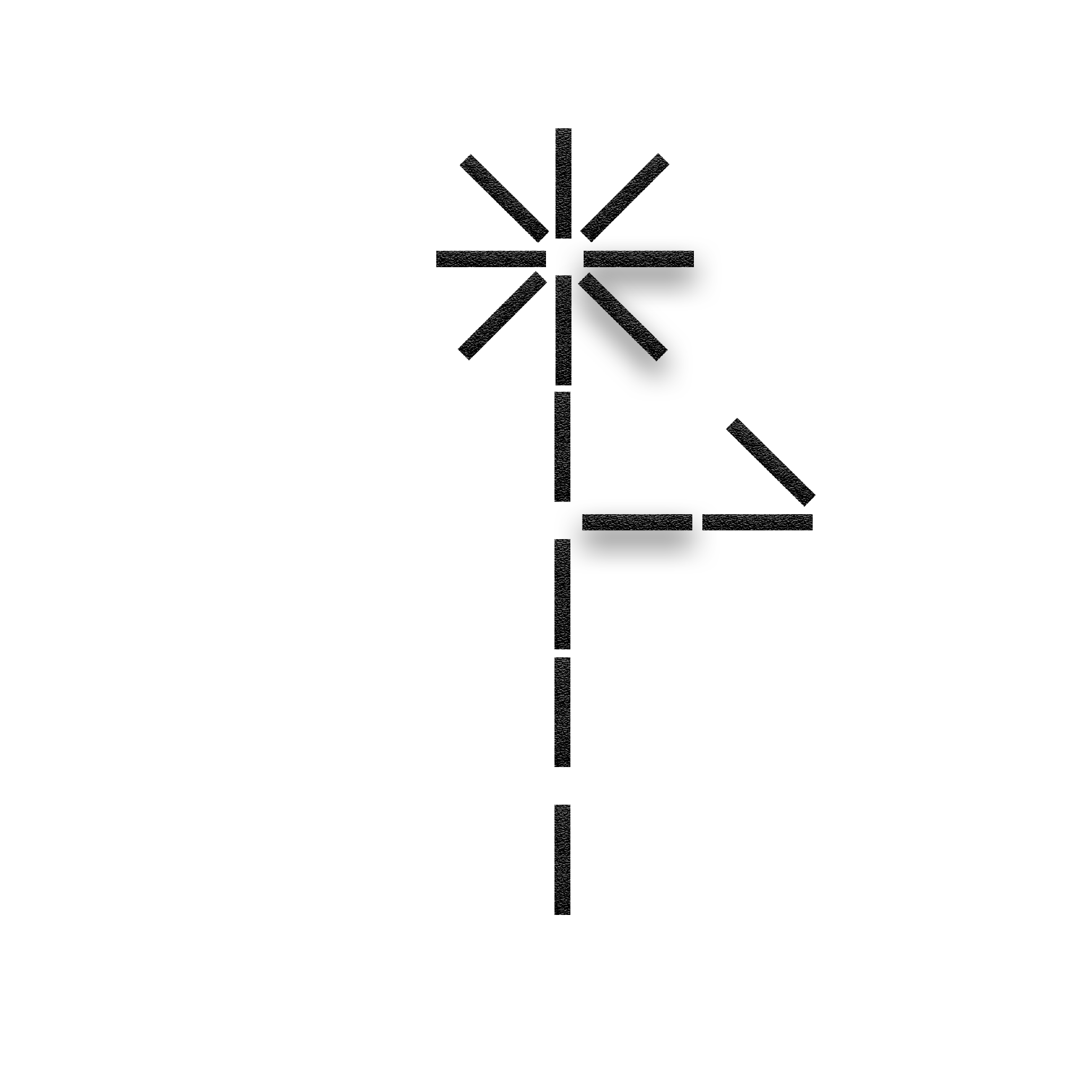}} & {\includegraphics[width=0.18\textwidth]{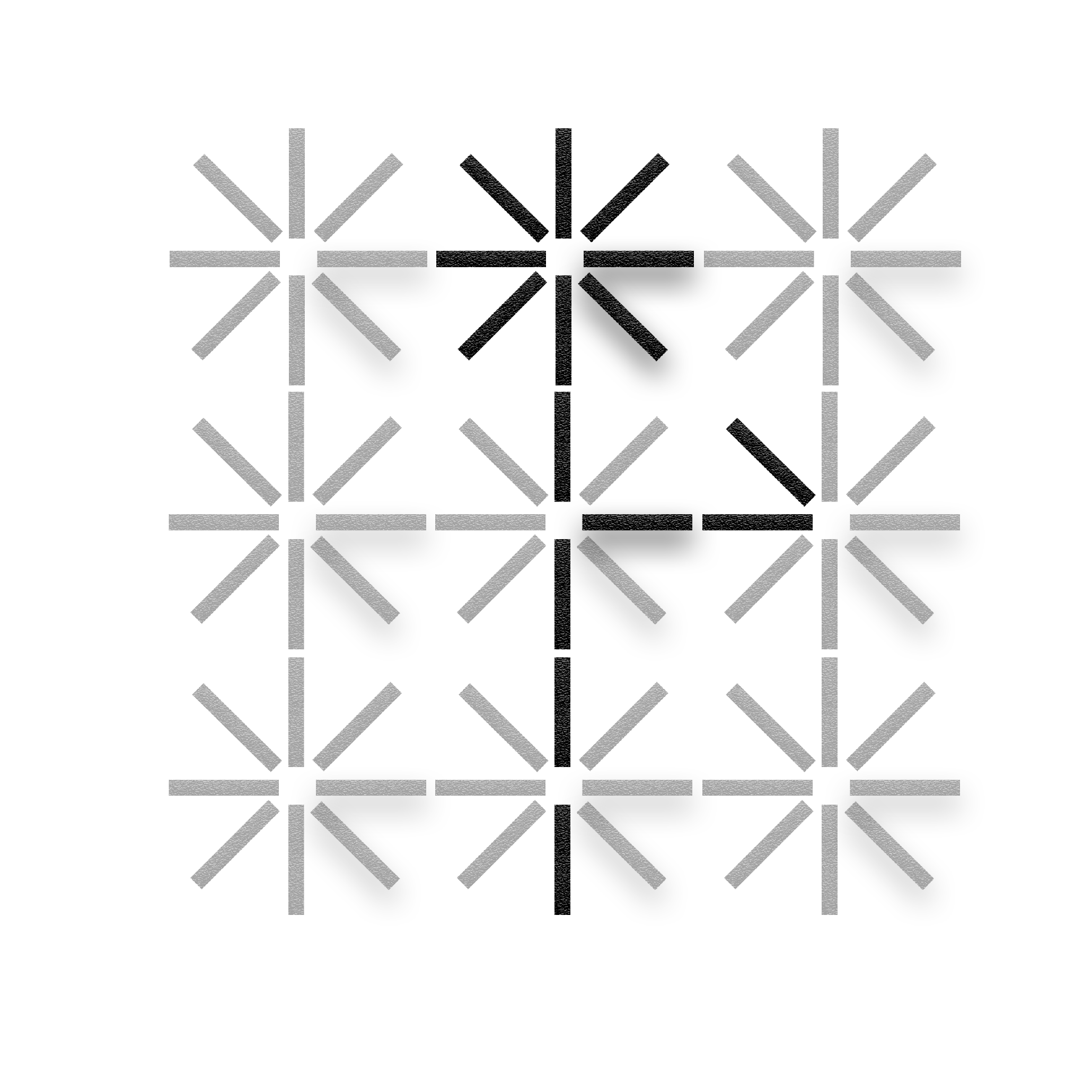}}\\
\hline
Eat meal/snacks &  {\includegraphics[width=0.18\textwidth]{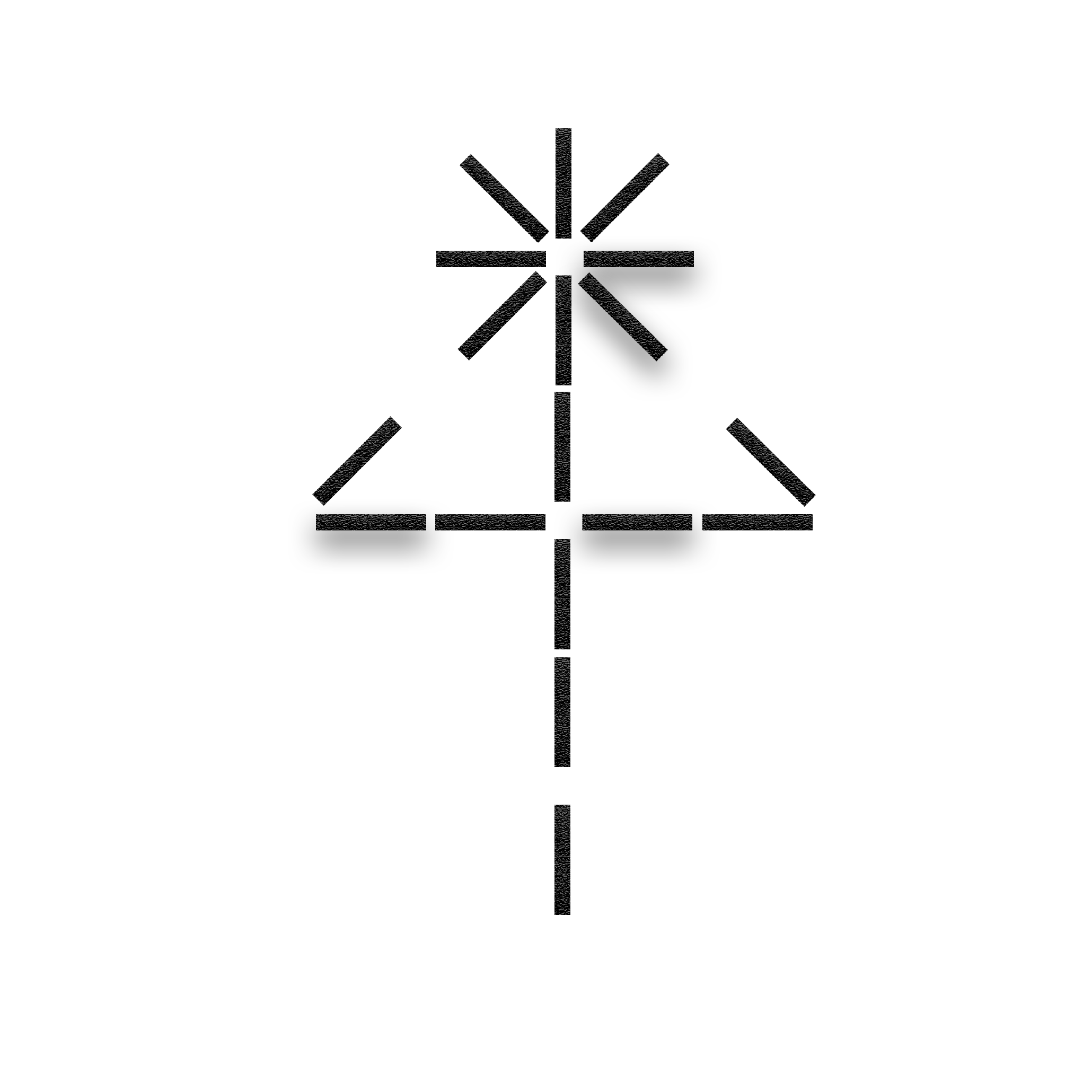}} & {\includegraphics[width=0.18\textwidth]{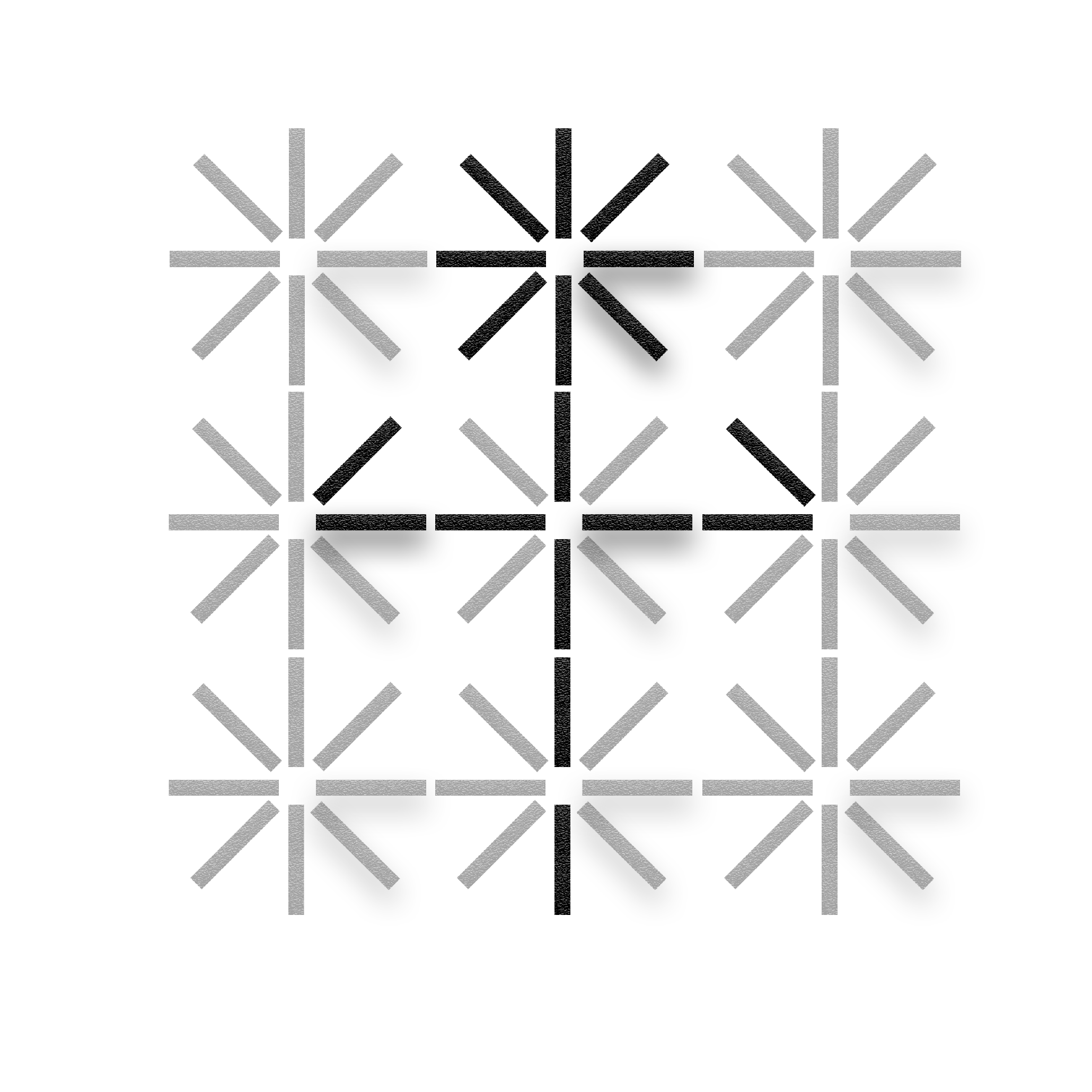}}\\
\hline
Hand-wave &  {\includegraphics[width=0.18\textwidth]{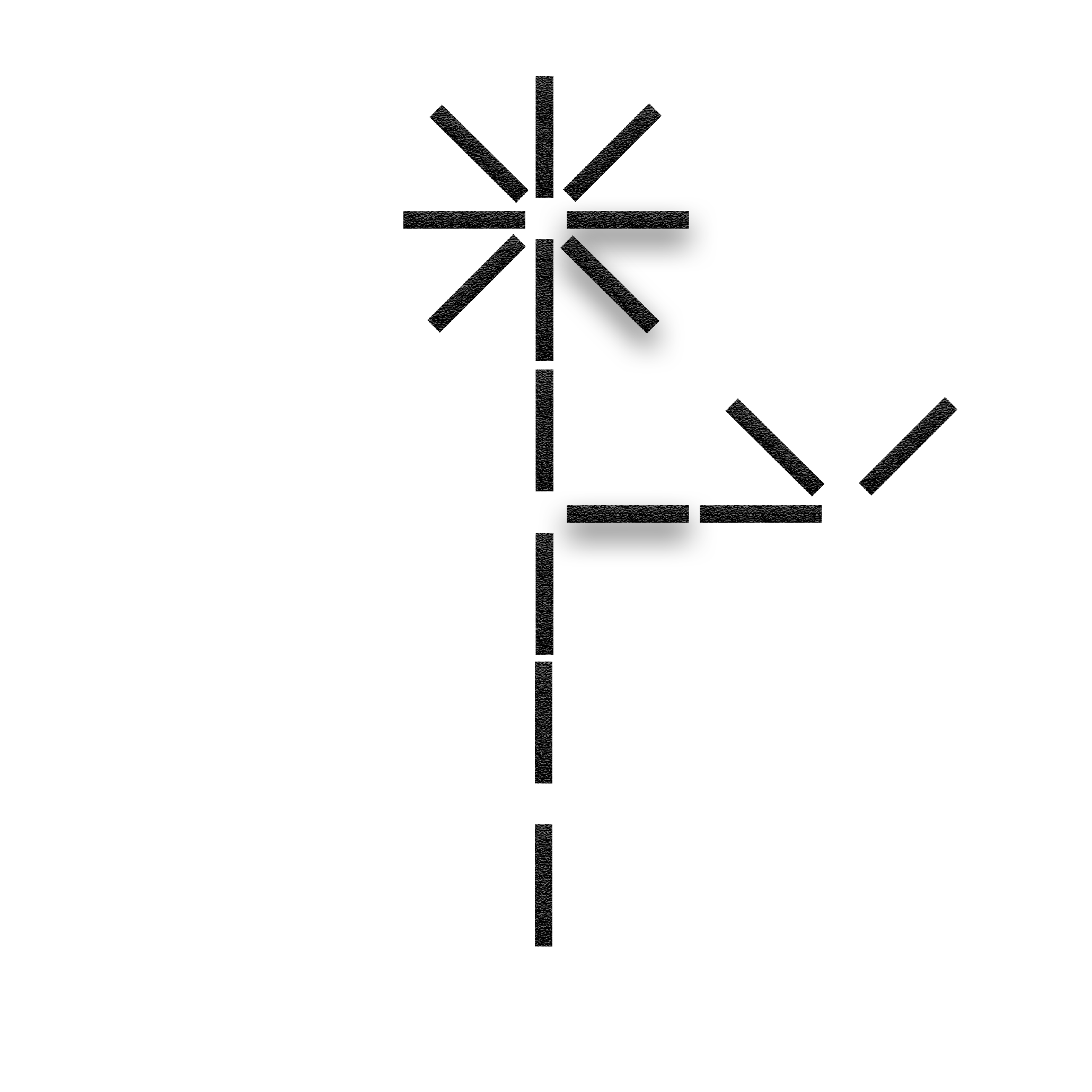}} & {\includegraphics[width=0.18\textwidth]{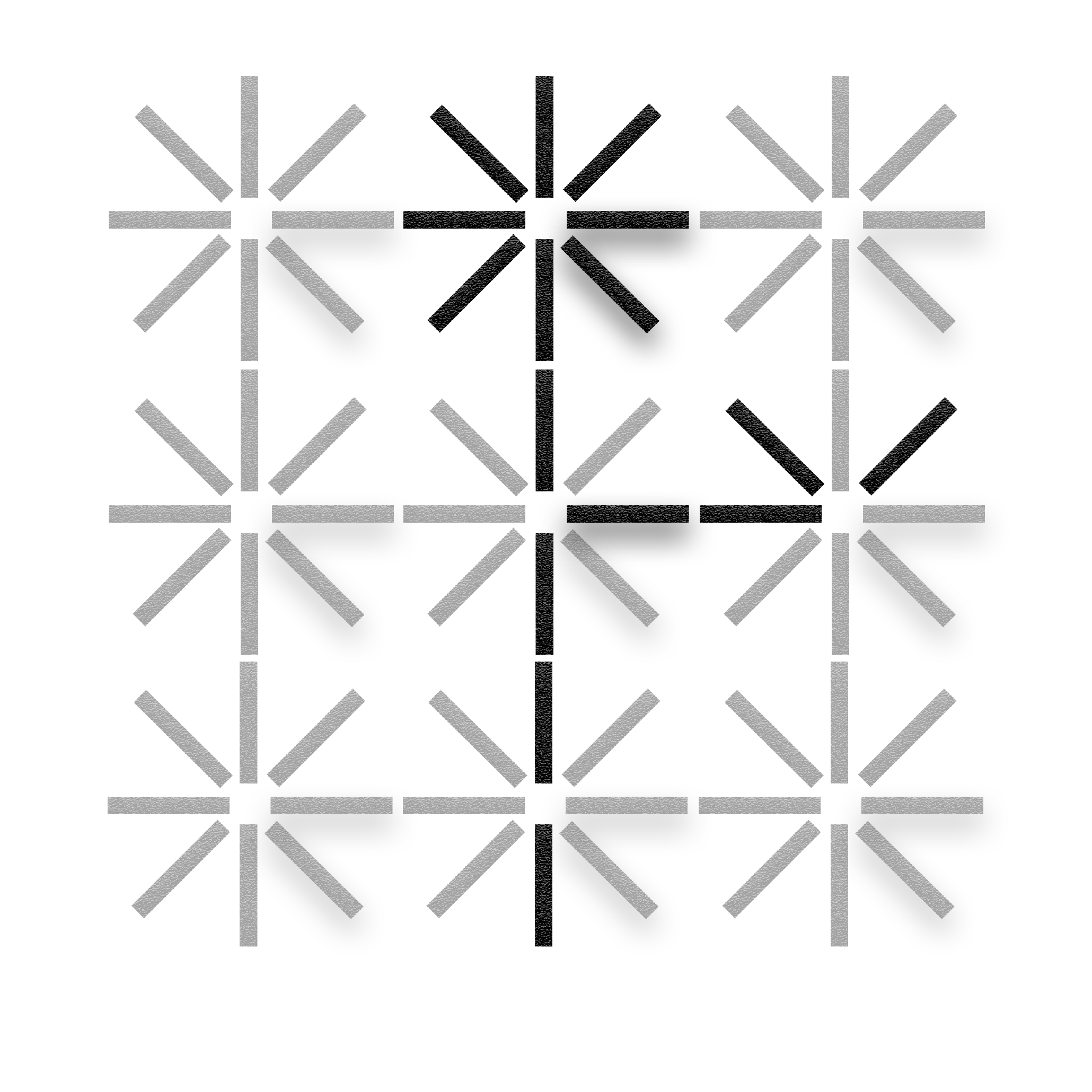}}\\
\hline
Kicking something &  {\includegraphics[width=0.18\textwidth]{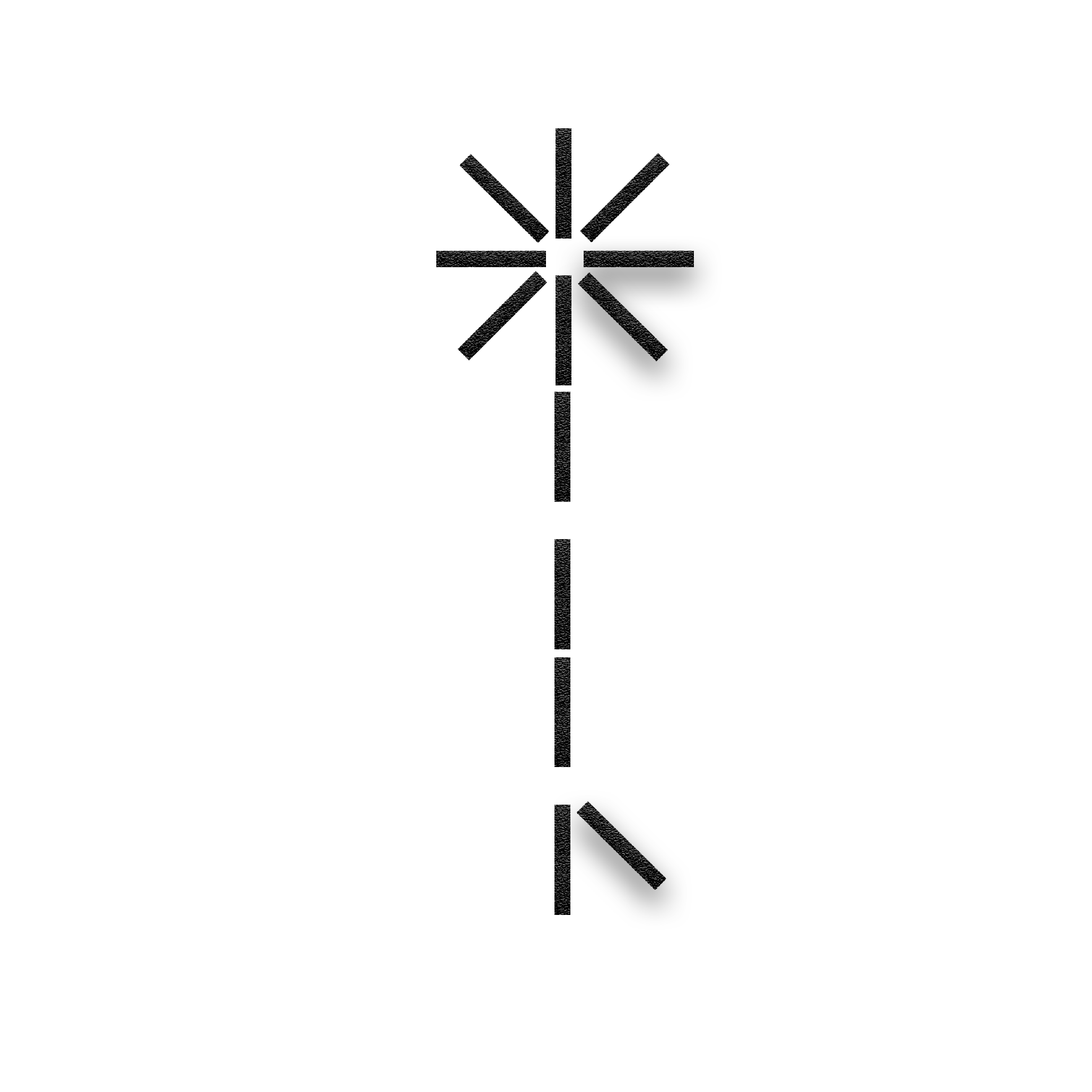}} & {\includegraphics[width=0.18\textwidth]{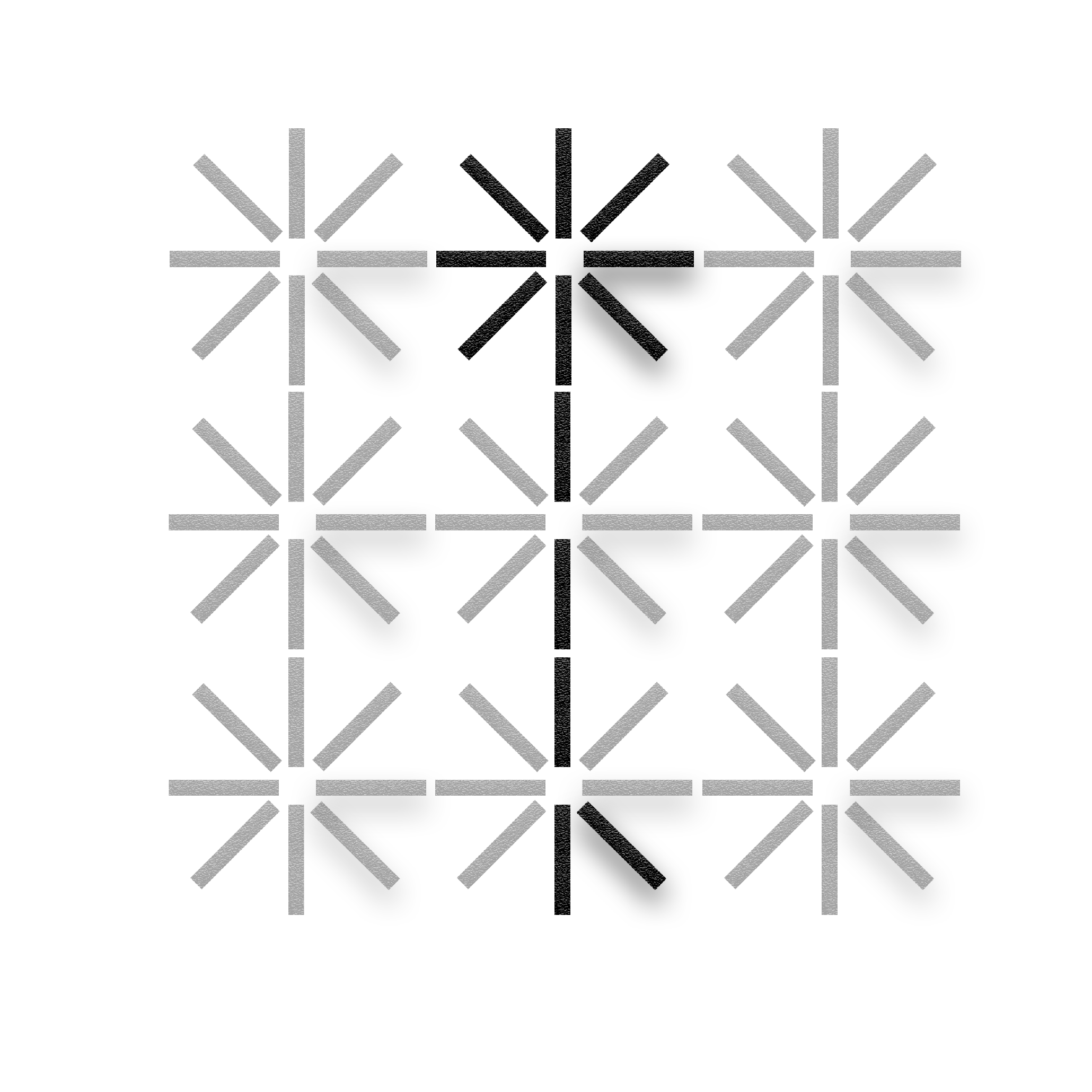}}\\
\hline
Touch head &  {\includegraphics[width=0.18\textwidth]{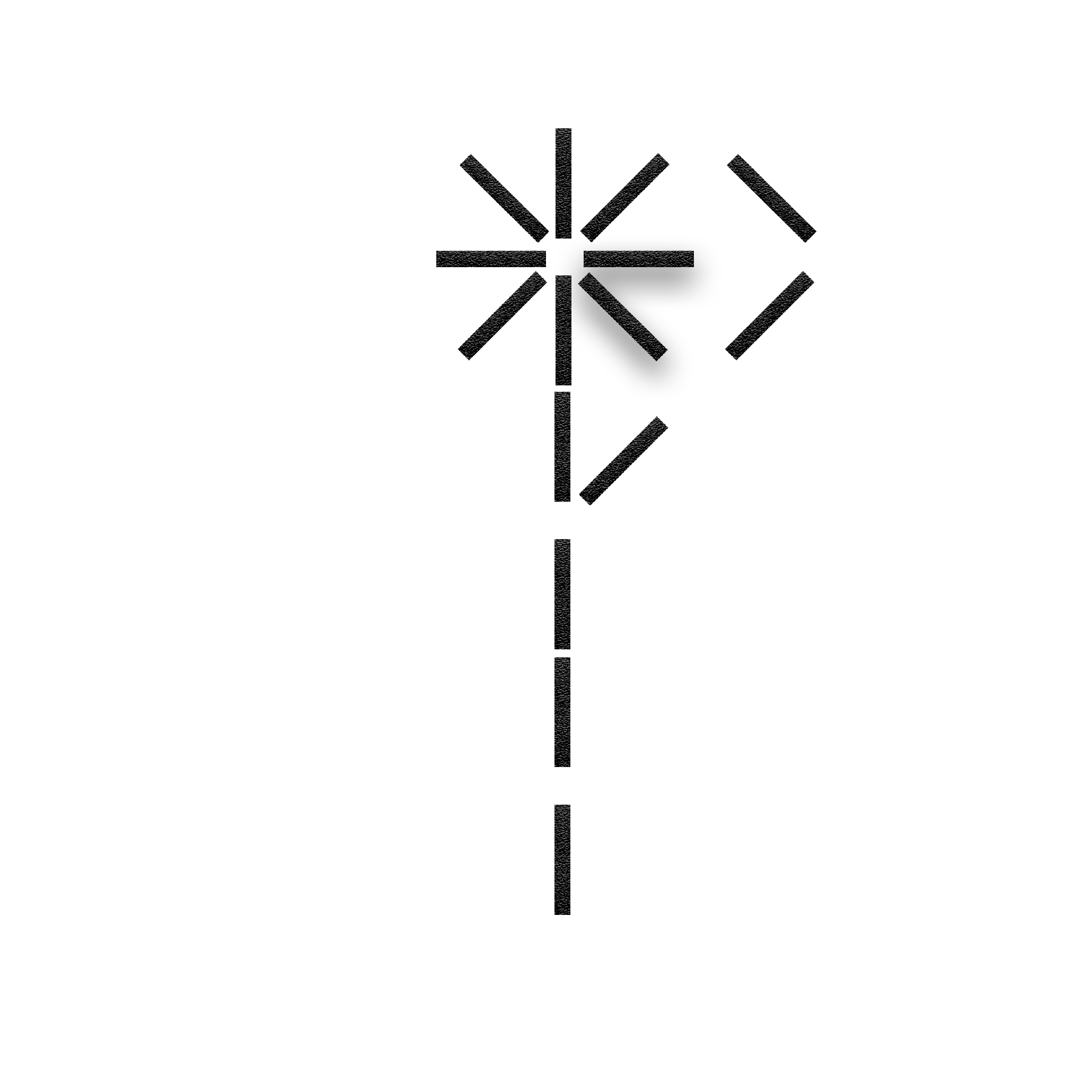}} & {\includegraphics[width=0.18\textwidth]{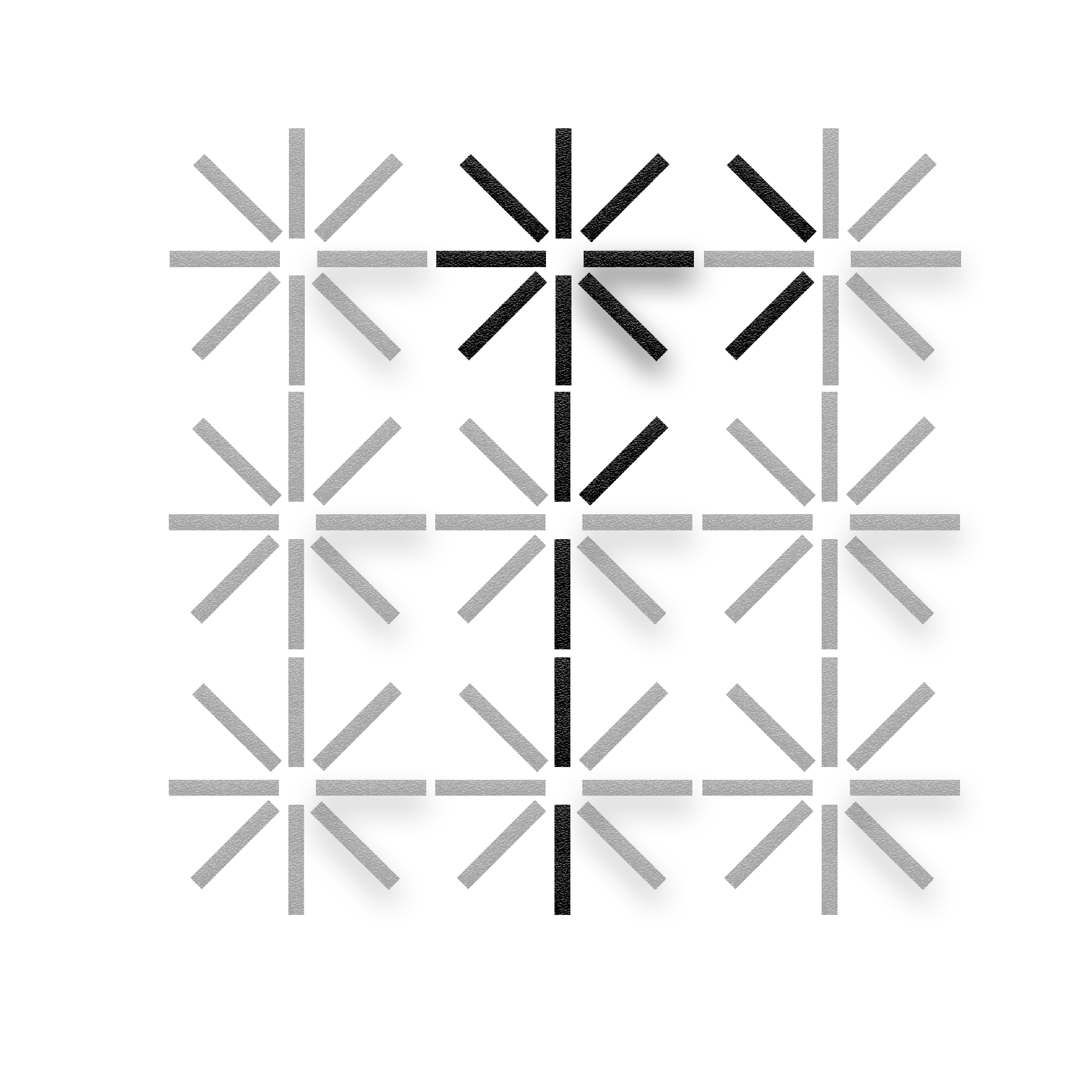}}\\
\hline
\end{tabular}
\end{center}
\end{table}

\section{Conclusion}
In this paper, we presented the obtained results using MS-G3D model for human action recognition on real scenes, in real-time. The obtained results showed some limitations of this model. In order to encounter these limitations and improve the recognition performance, we propose as future works to fuse this model with CNN using depth maps. In addition, we presented our proposed semantic labeling to represent human actions. Once an action is recognized, the corresponding label will be generated on the output device by raised cuboids. Our future works include two main tasks: - Improving the action recognition performance. – Providing more informative semantic labeling that will take into consideration face expressions.

\end{document}